\documentclass[conference]{IEEEtran}
\IEEEoverridecommandlockouts
\usepackage{cite}
\usepackage{amsmath,amssymb,amsfonts}
\usepackage{algorithmic}
\usepackage{graphicx}
\usepackage{textcomp}
\usepackage{xcolor}
\usepackage[caption=false]{subfig}
\usepackage[utf8]{inputenc}
\usepackage{booktabs}
\usepackage{url}
\usepackage{pifont}
\usepackage{tikz}

\def\BibTeX{{\rm B\kern-.05em{\sc i\kern-.025em b}\kern-.08em
    T\kern-.1667em\lower.7ex\hbox{E}\kern-.125emX}}
    
\newcommand\copyrighttext{%
  \footnotesize \textcopyright 2020 IEEE. Personal use of this material is permitted. Permission from IEEE must be obtained for all other uses, in any current or future media, including reprinting/republishing this material for advertising or promotional purposes, creating new collective works, for resale or redistribution to servers or lists, or reuse of any copyrighted component of this work in other works. Accepted to be Published in: Proceedings of the IJCNN 2020: International Joint Conference on Neural Networks, 19 - 24 July, 2020, Glasgow (UK)}
\newcommand\copyrightnotice{%
\begin{tikzpicture}[remember picture,overlay]
\node[anchor=south,yshift=10pt] at (current page.south) {\fbox{\parbox{\dimexpr\textwidth-\fboxsep-\fboxrule\relax}{\copyrighttext}}};
\end{tikzpicture}%
}

\begin{document}

\title{Towards Accurate Predictions and Causal `What-if' Analyses for Planning and Policy-making: A Case Study in Emergency Medical Services Demand}
\author{\IEEEauthorblockN{Kasun Bandara\IEEEauthorrefmark{1},
Christoph Bergmeir\IEEEauthorrefmark{1},
Sam Campbell\IEEEauthorrefmark{2},
Deborah Scott\IEEEauthorrefmark{2},
Dan Lubman\IEEEauthorrefmark{2}
}
\IEEEauthorblockA{\IEEEauthorrefmark{1}Faculty of Information Technology, Monash University, Melbourne, Australia.\\
herath.bandara@monash.edu, christoph.bergmeir@monash.edu}
\IEEEauthorblockA{\IEEEauthorrefmark{2} Turning Point, Eastern Health Clinical School, Monash University, Melbourne, Australia. \\
sam.campbell@monash.edu, debbie.scott@monash.edu, dan.lubman@monash.edu}}

\maketitle
\copyrightnotice

\begin{abstract}
Emergency Medical Services (EMS) demand load has become a considerable burden for many government authorities, and EMS demand is often an early indicator for stress in communities, a warning sign of emerging problems.
In this paper, we introduce Deep Planning and Policy Making Net (DeepPPMNet), a Long Short-Term Memory network based, global forecasting and inference framework to forecast the EMS demand, analyse causal relationships, and perform `what-if' analyses for policy-making across multiple local government areas. 
Unless traditional univariate forecasting techniques, the proposed method follows the global forecasting methodology, where a model is trained across all the available EMS demand time series to exploit the potential cross-series information available. 
DeepPPMNet also uses seasonal decomposition techniques, incorporated in two different training paradigms into the framework, to suit various characteristics of the EMS related time series data.
We then explore causal relationships using the notion of Granger Causality, where the global forecasting framework enables us to perform `what-if' analyses that could be used for the national policy-making process.
We empirically evaluate our method, using a set of EMS datasets related to alcohol, drug use and self-harm in Australia. The proposed framework is able to outperform many state-of-the-art techniques and achieve competitive results in terms of forecasting accuracy. We finally illustrate its use for policy-making in an example regarding alcohol outlet licenses.
\end{abstract}


\section{Introduction}
\label{sec:intro}

Emergency Medical Services (EMS), also known as ambulance services or paramedic services, provide out-of-hospital medical care and transportation to a medical facility for those in need of urgent medical attention. In general, minimising the response time to emergency calls is crucial for EMS to treat patients efficiently and effectively.

\begin{figure}[thbp]
\centerline{\includegraphics[width=0.45\textwidth, trim = 1cm 2cm 1cm 1cm]{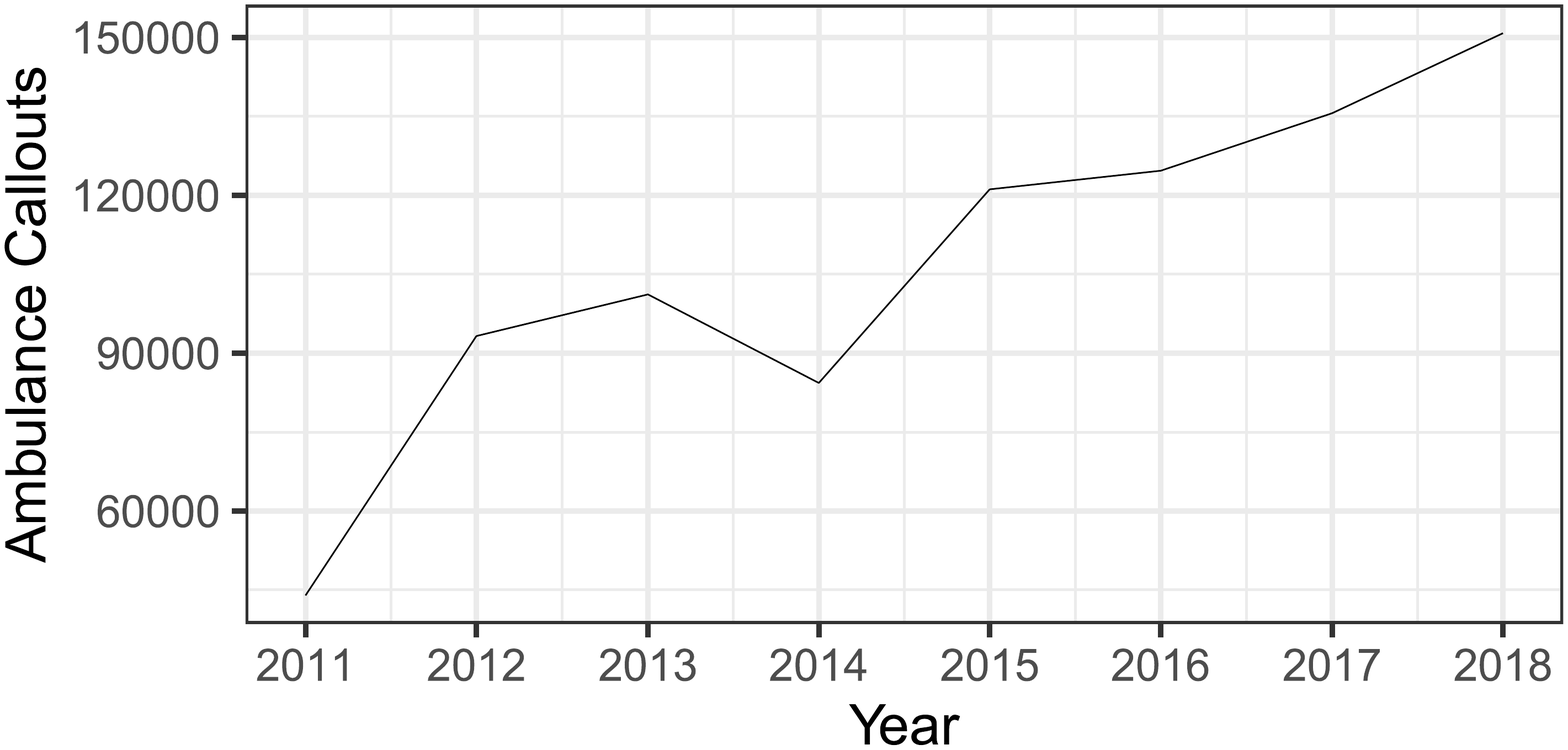}}
\caption{The reported number of ambulance attendances related to alcohol, drug use, self-harm, and mental health, for one of the states in Australia, over the past eight years.}
\label{demandtimeseries}
\end{figure}

Generating accurate and reliable spatio-temporal estimations of the future EMS demand is essential for authorities to proactively allocate the available resources, i.e., number of paramedics and size of ambulance fleets. A successful resource deployment strategy assures EMS deliver high-quality medical services in a timely manner. Furthermore, EMS demand patterns are often an early warning sign for certain problems in a community, and can therefore be useful for national-level policy makers, when introducing intervention strategies to address those problems.

The staggering growth of incidents requiring EMS assistance, combined with the scarcity of the resources have made EMS planning a challenging task for many countries \cite{Steins2019-eu,Zhou2015-cx,Setzler2009-km,Channouf2007-vw}. As a result, accurate EMS demand forecasting is becoming ever more important in the emergency services industry, so that the authorities can anticipate EMS requirements in advance to optimally utilise the available resources. Over the past two decades, numerous studies have been developed to improve the accuracy of EMS demand forecasts \cite{Steins2019-eu,Chen2016-ka,Zhou2015-cx,Wong2014-ok,Matteson2011-lv,Vile2012-bq,Setzler2009-km,Jones2009-gj,Channouf2007-vw}. However, many of the existing approaches hold strong prior distributional and structural assumptions about the EMS data. Furthermore, these methods are univariate forecasting models, which treat each time series separately, and forecast them in isolation. As a result, these models are unable to exploit the common EMS demand patterns present in collections of time series. Motivated by this need, we propose a forecasting framework, Deep Planning and Policy Making Net (DeepPPMNet), training a Long Short-Term Memory Network (LSTM) globally across different series, and therewith accounting for the cross-series information available in a set of EMS demand time series.

\begin{figure}[htbp]
\centerline{\includegraphics[width=0.60\textwidth, trim = 4cm 4cm 4cm 4cm]{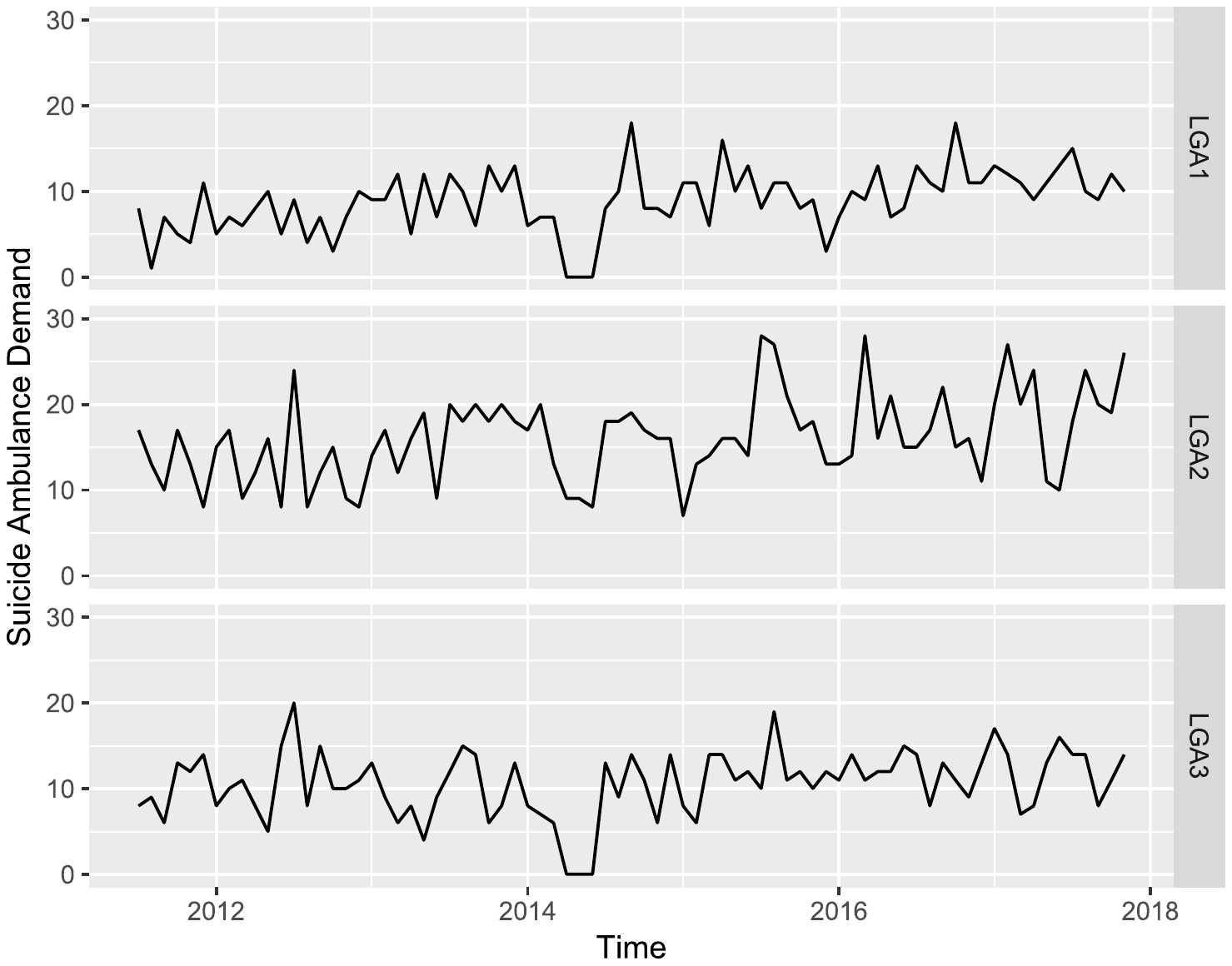}}
\caption{Monthly ambulance attendances for suicide attempts of three different LGAs for one of the states in Australia, over a time period of eight years.}
\label{lgademand}
\end{figure}

In this work, we use a case study of ambulance attendances related to alcohol, drug use, and self-harm in Australia. Similar to other developed nations, Australia is experiencing an increasing demand for EMS over the past few years. For example, Fig.~\ref{demandtimeseries} illustrates the growth of EMS demand related to alcohol, drug use, self-harm, and mental health, for one particular Australian state.
In the EMS capacity planning, generating sub-region demand forecasts has proven more useful than aggregated forecasts for an entire region \cite{Steins2019-eu,Zhou2015-cx,Setzler2009-km}. In Australia, states and territories are subdivided into statistical sub-regions known as Local Government Areas (LGAs). This breakdown enables authorities to deploy a fine-grained resource allocation strategy. Fig.~\ref{lgademand} illustrates such an example of EMS demand for suicide attempts, over the past eight years. Here, the EMS demand patterns of various LGAs can be similar and may share salient features in common, i.e., similar ambulance demand peaks and drops. In order to exploit these similar EMS demand patterns, the proposed DeepPPMNet framework employs a global forecasting methodology (GFM) that simultaneously learns from a collection of time series \cite{Januschowski2020-ud}. 

We subsequently perform causal inference on the potential factors that influence the EMS demand, employing the concept of Granger Causality (GC)~\cite{Granger1980-xw,Seth2007-on}, that we define according to Granger~\cite{Granger1980-xw,Seth2007-on} as follows. 
Given a target time series $Y_t$, and (multivariate) external factors $W_t$ and $Z_t$. 
Let us assume that we have a predictive model that uses past values of $Y_t$ and $W_t$ to predict $Y_{t+1}$. We now define that $Z_t$ Granger causes $Y_{t+1}$, if we are able to improve the accuracy of predicting $Y_{t+1}$ by including past values of $Z_t$ as exogenous values in our model.
Granger~\cite{Granger1980-xw} notes that causality is a complicated concept and no universally accepted definition exists. Granger causality assumes that a cause has to precede the effect in time, i.e., the future cannot cause the past. Then, in the hypothetical case in which $W_t$ and $Z_t$ together cover all possible factors, Granger argues that Granger causality is true causality. In practice, the strength of Granger causality depends heavily on the careful analysis and inclusion of possible influencing factors $W_t$, which effectively determines if Granger causality measures rather a correlation or suggests true causal relationships.

The GC analysis subsequently allows us to perform `what-if' analyses, in the sense that if we are confident that we have identified exogenous variables $Z_t$ as causal factors, we can deliberately change their values to analyse how the predicted outcomes change accordingly. We note that using a GFM based forecasting model like DeepPPMNet greatly improves this methodology, as the model is trained across all the variations that exist among a collection of time series, which makes the `what-if' analysis feasible even for relatively constant features or feature values that have not been observed in a particular time series before.
The main contributions of our paper are as follows:

\begin{itemize}
\item To the best of our knowledge, our work is the first study that employs a GFM in a collection of EMS demand time series for forecasting. We evaluate our model on a real-world EMS demand dataset, which consists of ambulance attendances related to alcohol overdose, suicide attempts, and other drug related harms reported in Australia, and show that our framework achieves competitive forecasting accuracy.
\item Building on these results, we then show how DeepPPMNet provides a platform to conduct GC-based `what-if' analyses, and allows decision makers to assess the factors that could drive the EMS demand, for which they can implement timely intervention or prevention strategies.
\end{itemize}

The remainder of this paper is organized as follows.  In Section~\ref{sec:relatedwork}, we review the state of the art in this domain, and discuss the advent of global forecasting models. In Section~\ref{sec:design}, we discuss the proposed DeepPPMNet framework in detail. Our experimental setup is presented in Section~\ref{sec:experiments}, where we apply the proposed DeepPPMNet framework to multiple EMS demand datasets extracted from an internationally unique dataset of coded EMS clinical records. Finally, Section~\ref{sec:conclusion} concludes the paper.

\section{Prior Work}
\label{sec:relatedwork}

Early studies to forecast EMS demand are mostly based on exponential smoothing models \cite{Baker1986-vk}, autoregressive moving average models \cite {Channouf2007-vw,Wong2014-ok}, factor models \cite{Matteson2011-lv}, and simple averaging models \cite{Brown2007-ij}. Baker and Fitzpatrick \cite{Baker1986-vk} use a Holt-Winter's exponential smoothing model along with goal programming to forecast EMS demand in South Carolina. Channouf et al. \cite{Channouf2007-vw} investigate the use of ARIMA models with special-day effects to predict EMS calls in Calgary, Alberta. Furthermore, Wong et al. \cite{Wong2014-ok} use ARIMA models to measure the effect of weather factors for daily ambulance call out demand. Also, Brown et al. \cite{Brown2007-ij} use various forms of averaging techniques to forecast the hourly emergency call volumes. Matteson et al. \cite{Matteson2011-lv} propose a combined inter-valued time series model with a dynamic latent factor structure using hourly emergency call demand data from Toronto, Canada.

Overcoming the limitations of traditional EMS prediction approaches, Setzler et al. \cite{Setzler2009-km} introduce a neural network (NN) based methodology to generate spatio-temporal forecasts for an EMS agency in North Carolina. Based on demand data provided by the Welsh Ambulance Service Trust, Vile et al. \cite{Vile2012-bq} evaluate the application of singular spectrum analysis for daily EMS demand forecasting. Zhou and Matteson \cite{Zhou2015-cx} develop a spatio-temporal kernel density estimation approach to predict ambulance demand in Toronto, Canada. More recently, Chen et al. \cite{Chen2016-ka} use a combination of machine learning techniques to forecast the pre-hospital emergency demand in New Taipei city, while Steins et al.~\cite{Steins2019-eu} use a zero-inflated poisson regression model to forecast ambulance call demand in Sweden.

In parallel to these developments, deep learning based applications have achieved state-of-the-art performance in many domains \cite{Smyl2019-cb,Bandara2019-ua,Flunkert2017-mt}. In the time series forecasting field, the modern advances are mainly around deep learning based GFMs. These models are becoming a strong alternative to the univariate time series forecasting models, in situations where a collection of related time series is available. This is particularly evident from the recent success of such models in forecasting competitions \cite{Smyl2019-cb} and the use in industry applications \cite{Bandara2019-ua,Flunkert2017-mt}. As highlighted in Section~\ref{sec:intro}, compared to univariate time series models, which rely only on the individual level time series data, GFM models benefit from training across multiple time series, allowing them to learn key patterns available in a collection of time series. 

\section{DeepPPMNet Framework}
\label{sec:design}
DeepPPMNet is a GFM based forecasting framework, which is designed to train across multiple time series. The DeepPPMNet consist of three layers, namely: 1) the pre-processing layer, 2) the DeepPPMNet training layer, and 3) the post-processing layer. In the following, we discuss the functionality of these layers in detail.

\subsection{Pre-processing layer}
\label{sec:preprocessing}

The pre-processing layer of DeepPPMNet is a three-stepped process. First, the set of EMS demand time series are normalised using a \textit{meanscale} transformation strategy to scale various ambulance call volume ranges of multiple LGAs~\cite{Flunkert2017-mt,Bandara2019-bg}, as DeepPPMNet is trained across a group of time series.

In the second step of pre-processing layer, following the suggestions of \cite{Hewamalage2019-il,Bandara2019-iv}, we perform a log transformation strategy to stabilise the variance of the EMS demand time series. In addition to stabilising the variance, the log transformation also assures additive seasonality and trend in the series, which is necessary for our decomposition layer, which is the last step of our pre-processing layer.

We use a time series decomposition layer as the final pre-processing step of DeepPPMNet. The main objective of this layer is to seasonally adjust the time series, i.e., removing the seasonal component from a time series, and thereby supplementing the subsequent DeepPPMNet learning phase. This deseasonalisation strategy is advocated by many studies, when using neural networks for time series forecasting~\cite{Bandara2019-si,Bandara2019-iv}. Due to the promising performance of \textit{Prophet} \cite{Taylor2017-lw} as a multi-seasonal decomposition technique in the recent past~\cite{Bandara2019-si}, we use Prophet as an additive, single-seasonal decomposition technique to obtain the seasonal components present in a time series. This additive decomposition can be defined as follows:

\begin{equation}
X_t = \hat{S}_t + \hat{T}_t  + \hat{R}_t + \hat{H}_t
\label{multiadditive}
\end{equation}
 
Here, $X_t$ denotes the EMS demand observation at time $t$, and the decomposed seasonal, trend, remainder, and holiday effect components of the time series are represented by $\hat{S}_t $, $\hat{T}_t$, $\hat{R}_t$, and $\hat{H}_t$ respectively. We perform the seasonal extraction, after fitting the Prophet model using the \verb|prophet| package implemented in R \cite{Taylor2018-wb}. Also, this deseasonalisation process can be replaced with other additive decomposition techniques such as seasonal and trend decomposition using loess (STL) \cite{Cleveland1990-rc}.

\subsection{DeepPPMNet Training layer}
As the primary prediction unit of DeepPPMNet, we use an LSTM \cite{Hochreiter1991-en}, a special type of neural network which is naturally suited for modelling time series data, because of its ability to capture long-term dependencies in sequences. LSTMs have been heavily used in sequence modelling tasks, such as Natural Language Processing, speech recognition, and more recently have received significant attention in the time series forecasting domain \cite{Smyl2019-cb,Bandara2019-iv,Flunkert2017-mt}. It is also worth mentioning that the proposed DeepPPMNet framework can potentially be employed with other RNN variants such as the Elman RNN \cite{Elman1990-my} or the Gated Recurrent Unit (GRU) cell \cite{Cho2014-or}. For more discussions on the LSTM architecture, we refer to Bandara et al.~\cite{Bandara2019-iv}.

\subsubsection{Moving Window Transformation}
In order to train the DeepPPMNet, we use the past observations of time series ($X_i$) in the form of input and output frames. Following recommendations in the literature~\cite{Smyl2019-cb,Bandara2019-iv,Hewamalage2019-il}, we achieve this by applying the Moving Window (MW) tranformation strategy to each pre-processed time series. In this process, the size of the output frame is equivalent to the size of the intended forecasting horizon $M$, i.e., a Multi-Input Multi-Output (MIMO) principle is used in multi-step forecasting. This enables the DeepPPMNet to directly predict all the future observations up to the intended forecasting horizon $X^M_{i}$, while avoiding prediction error accumulation at each forecasting step \cite{Ben_Taieb2012-re}. For further discussions of the MW transformation strategy we refer to Hewamalage et al.~\cite{Hewamalage2019-il}.

\subsubsection{DeepPPMNet Training Paradigms}
\label{sec:trainingpara}
Following Bandara et al.~\cite{Bandara2019-si}, we apply two training paradigms to the output generated by the decomposition layer. 

In the first training paradigm, Deseasonalised Approach (DS), the MW strategy is applied to a seasonally adjusted time series, and the transformed input and output windows are used to train the DeepPPMNet. As the seasonal components are excluded from the time series, an additional reseasonalisation process is introduced in the post-processing layer to predict the future seasonal values of the time series. These seasonal values are estimated by repeating the last seasonal components of the time series (extracted by Prophet) to the intended forecast horizon. The DS approach can be seen as a `boosting' ensemble technique \cite{Schapire2003-xb}, where the Prophet method used for the deseasonalisation process acts as a weak base learner to supplement the LSTM training procedure.

In the second training paradigm, Seasonal Exogenous Approach (SE), instead of detaching the seasonal components from the time series, we use the extracted seasonal components as exogenous variables together with the original time series (after normalising and log-transforming). As the original distribution of the time series is used to generate the training windows (using the MW process), the DeepPPMNet is expected to forecast all components of a time series, including extrapolation of the seasonality. In this way, SE supplements the DeepPPMNet by assisting to model the seasonal trajectories of a time series.

\begin{figure*}[htp]
\subfloat[The proposed DS training paradigm used in the DeepPPMNet framework]{%
  \centerline{\includegraphics[width=0.45\textwidth, trim = 8cm 8cm 7cm 8cm]{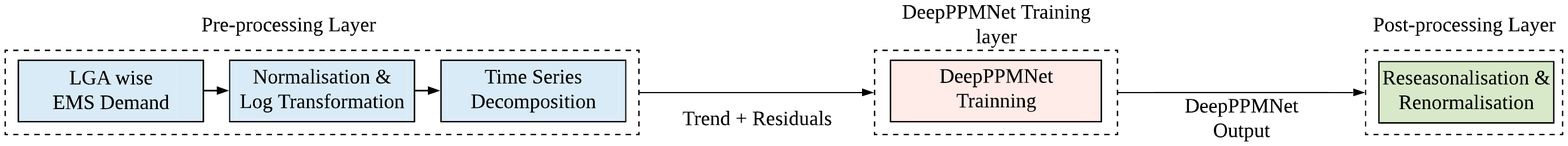}}
}
\vspace{1mm}
\subfloat[The proposed SE training paradigm used in the DeepPPMNet framework]{%
  \centerline{\includegraphics[width=0.45\textwidth, trim = 8cm 8cm 7cm 8cm]{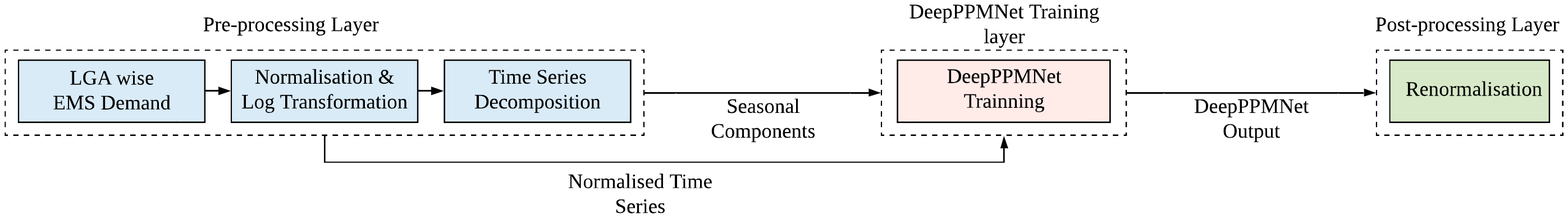}}
}
\caption{An overview of the proposed DeepPPMNet framework. A reseasonalisation phase is required for the post-processing layer of the DS approach, as the seasonal components are not included in the DeepPPMNet training phase. Whereas, the SE approach uses the pre-processed time series along with the seasonal components as external variables, and directly forecast all the components of the time series, thus a reseasonalisation phase is not required.}
\label{lstm-schemes}
\end{figure*}

Fig.~\ref{lstm-schemes} illustrates the overview of the proposed two training paradigms and the overall procedure of the DeepEMSFNet .

\subsubsection{DeepPPMNet Learning Scheme}

Following the recommendations of Bandara et al.~\cite{Bandara2019-iv} and Hewamalage et al.~\cite{Hewamalage2019-il}, we implement the ``LSTM Stacking Layer'' design pattern in DeepPPMNet. Here, a stacking of LSTM layers is followed by an affine neural layer (excluding the bias component) to map each LSTM cell output to the dimension of the intended forecasting horizon. As we follow the MIMO approach, when generating the input and output windows, the dimension of this fully connected neural layer is equivalent to the size of the output window. 

To avoid possible network saturation effects that occur in neural networks \cite{Bandara2019-iv,Hewamalage2019-il}, we perform a local normalisation process on the training windows, before using them for the DeepPPMNet training procedure. We implement two different normalisation strategies for each proposed training paradigm. The trend component of the last value of the input window is used in the DS approach, whereas in the SE approach, the mean value of each input window is used. 

We implement the above training paradigms and learning schemes of the DeepPPMNet using TensorFlow, an open-source deep-learning toolkit \cite{Abadi2016-rr}. The operating environment in EMS is often volatile and may present irregular demand patterns and anomalies. To be more robust under these circumstances, we use L1-norm as the primary learning objective function of DeepPPMNet. Moreover, we also include a L2-regularisation term to minimise possible overfitting of the network.

\subsection{Post-processing Layer}

The post-processing layer of DeepPPMNet consists of two phases, namely the reseasonalisation phase and the renormalisation phase. Firstly, in the reseasonalisation phase, the seasonal components estimated by the Prophet method are added back to the forecasts generated by the LSTM. As discussed in Section~\ref{sec:trainingpara}, this phase is only required by the DS training paradigm, as the seasonal components of the time series are not included in the DeepPPMNet training procedure. Secondly, in the renormalisation phase, the corresponding forecasts are back-transformed into the original scale, by adding back the local normalisation factor used in each training window, taking the exponent of the values, and finally multiplying by the mean scaling factor of each time series that is used in the normalisation process. The renormalisation phase is required by both the DS and SE training paradigms, as the mean normalisation and log transformation are common processes for both training paradigms.

\section{Experiments}
\label{sec:experiments}

In this section, we present the experimental setup used to evaluate the DeepPPMNet framework on real-world EMS datasets. This includes a summary of the datasets, error metrics, hyper-parameter selection, benchmarks and a discussion of the results obtained.

\subsection{Datasets}

We use an internationally unique, national dataset of coded ambulance clinical records held by Turning Point, an Australian addiction research and education centre. The dataset (The National Ambulance Surveillance System or NASS) \cite{TurningPoint2020-nw} holds surveillance data on alcohol and other drug, self-harm and mental health-related ambulance attendances across 5 of the 6 Australian states and 2 territories. We subdivide the dataset into five categories of ambulance attendance datasets namely: 1) the alcohol overdose (AO) dataset, 2) the suicide-attempt (SA) dataset, 3) the heroin overdose (HO) dataset, 4) the methamphetamine overdose (MO) dataset, and 5) the prescription Opioids overdose (PO) dataset. For further definitions of these categories, we refer to \cite{Lloyd2013-ky}.

We use 8 years worth of monthly EMS data available from September-2011 to May-2019. The last 12 months of EMS observations, i.e., May-2018 to May-2019, are reserved for testing. The size of the training output window is 12, as this is the intended forecasting horizon. Also, we choose the size of the training input window as 15 (12*1.25), following the heuristic proposed in \cite{Hewamalage2019-il, Bandara2019-ua}. The number of LGAs, i.e., number of time series, present in the AO, SA, HO, MO and PO categories are 89, 88, 80, 87 and 87, respectively.

\begin{figure}[htbp]
\centerline{\includegraphics[width=0.60\textwidth, trim = 2cm 2cm 2cm 2cm]{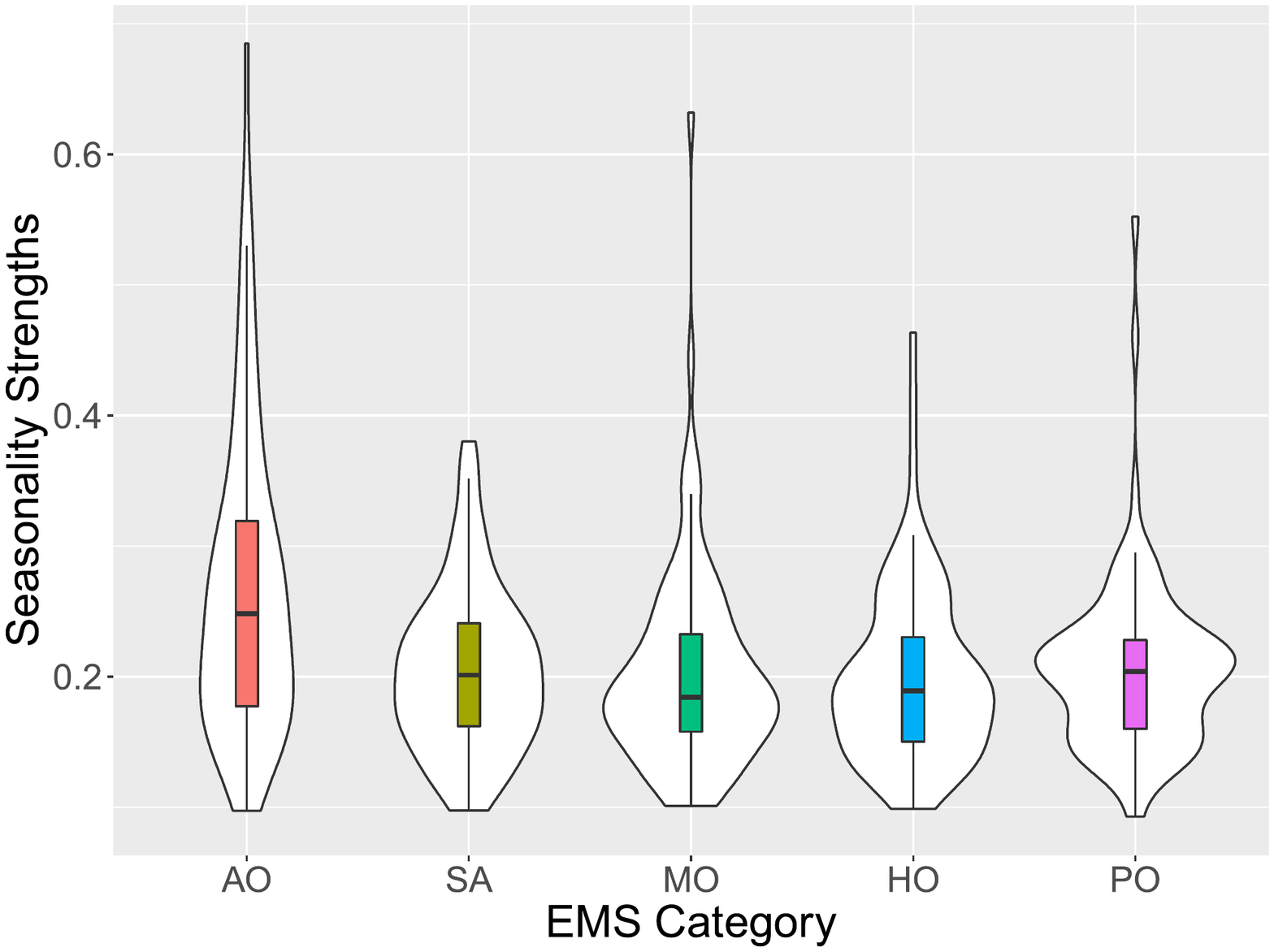}}
\caption{The yearly seasonal strength distributions of the AO, SA, HO, MO and PO categories. The seasonal components are extracted using Prophet as a decomposition technique.}
\label{lgademand2}
\end{figure}

Fig.~\ref{lgademand2} shows violin plots generated using the seasonal components of each EMS category. This enables us to analyse how seasonal strength can affect the performance of the DeepPPMNet training paradigms, i.e., DS and SE approaches. We observe that the AO category carries a higher seasonal strength compared to other categories. This can be mainly attributed to the higher consumption of alcohol during the holiday periods.

\subsection{Error Metrics}

To assess the accuracy of the DeepPPMNet, we use two evaluation metrics commonly used in the time series forecasting literature, namely the symmetric Mean Absolute Percentage Error (sMAPE) and the Mean Absolute Scaled Error (MASE), which are scale independent error measures \cite{Hyndman2006-ue}. 
Using these measures to calculate the forecast errors independently of the scale of the time series leads to effectively every series contributing equally to the error, even though the LGA-wise EMS dataset may contain different scales of time series. This may be appropriate for certain applications but not for others. However, these measures are also commonly used in forecasting competitions nowadays, and can be considered as standard evaluation measures in forecasting. The sMAPE and MASE error measures are defined as follows:

\begin{equation}
\text{sMAPE} = \frac{2}{m}\sum_{t=1}^{m}\left(\frac{\left|F_t - Y_t\right|}{\left| F_t\right| + \left| Y_t\right|} \right)
\label{smape}
\end{equation}

\begin{equation}
\text{MASE} = \frac{1}{m}\frac{\sum_{t=1}^{m}|F_t - Y_t|}{\frac{1}{n-S}\sum_{t=S+1}^{n}|Y_{t}-Y_{t-S}|}
\label{mase}
\end{equation}

Here, $F_t$ denotes the forecasts generated by DeepPPMNet, and $Y_t$ is the actual EMS observation at time $t$. Furthermore, $m$ represents the number of data points in the test set and $n$ is the number of observations in the training set of a time series. $S$ refers to the frequency of the seasonality in a given time series, set to 12 in our experiments, as we use monthly EMS demand time series. The mean and median of these error measures are reported in our evaluation.

\subsection{Hyper-parameter Tuning and Optimisation}

A Python implementation of a Bayesian hyper-parameter optimisation process, the sequential model-based algorithm configuration (SMAC) \cite{Hutter2011-wa} is used to autonomously determine the optimal hyper-parameters of DeepPPMNet. As we employ COntinuous COin Betting \cite{Orabona2017-cd}, tuning the learning rate is not necessary in DeepPPMNet. Table \ref{tab:parametergrid} summarises the bounds of the hyper-parameter values used in the DeepPPMNet learning process.

We also address the parameter uncertainty of our proposed framework by training DeepPPMNet on 10 different Tensorflow graph seeds, using the optimised hyperparameters from SMAC. We generate the final forecasts of DeepPPMNet by computing the median of the forecasts that are produced from different initialisation seeds of DeepPPMNet.

\begin{table}
\caption{DeepPPMNet hyper-parameter grid}
\begin{center}
\begin{tabular}{lcc}
	\toprule
	Model Parameter            			&Minimum value 			&Maximum value\\ \hline
	LSTM-cell-dimension  				&20						&50	\\
	Mini-batch-size						&1						&10	\\
	Epoch-size							&2						&10 \\
	Maximum-epochs						&3						&40	\\
	Hidden Layers						&1						&2\\
	Gaussian-noise-injection			&$10^{-4}$				&$8 \cdot 10^{-4}$\\
	Random-Normal-Initialiser			&$10^{-4}$				&$8 \cdot 10^{-4}$\\
	L2-regularisation-weight			&$10^{-4}$				&$8 \cdot 10^{-4}$\\ \hline
\end{tabular}
\label{tab:parametergrid}
\end{center}
\end{table}

\subsection{Benchmarks and DeepPPMNet variants}
We use a collection of state-of-the-art univariate forecasting techniques to benchmark against. This includes well-established methods for monthly time series forecasting, such as \textit{ETS} \cite{Hyndman2008-yd}, \textit{ARIMA} \cite{Box2015-bz}, and recently developed forecasting methods, such as \textit{BaggedETS} \cite{Bergmeir2016-zk}, and \textit{Prophet} \cite{Taylor2017-lw}. Here,  the ETS, ARIMA, and BaggedETS methods are used from the \verb|forecast| package implemented in R \cite{Hyndman2015-vm}, while the Prophet method is used from the \verb|prophet| package in R.

When building GFMs, as the first approach, we train separate DeepPPMNet models on each EMS category (GFM-CAT). As the second approach, we build one GFM across all the available EMS category datasets (GFM-ALL). Based on the proposed training paradigms and the GFM modelling approaches, we define the following variants of DeepPPMNet.

\begin{itemize}
\item \textit{DeepPPMNet-DS}: The DeepPPMNet framework that uses DS as the training paradigm and a separate GFM is built on each EMS category.
\item \textit{DeepPPMNet-SE}: The DeepPPMNet framework that uses SE as the training paradigm and a separate GFM is built on each EMS category.
\item \textit{DeepPPMNet-DS-ALL}: The DeepPPMNet framework that uses DS as the training paradigm and one GFM is trained across all the EMS categories.
\item \textit{DeepPPMNet-SE-ALL}: The DeepPPMNet framework that uses SE as the training paradigm and one GFM is trained across all the EMS categories.
\end{itemize}

\subsection{Results and Discussion}
Table~\ref{tab:ao} shows the evaluation summary of the DeepPPMNet variants and benchmarks for the 89 LGAs of the AO category, ordered by the Mean sMAPE. For each column, the results of the best performing methods are marked in boldface.  According to Table~\ref{tab:ao}, we see that the proposed DeepPPMNet-DS-ALL variant achieves the best Mean sMAPE, Mean MASE, and Median sMAPE, while Prophet obtains the best Median MASE. Also, among the proposed variants, we observe that the DeepPPMNet variants with the DS training paradigm achieve better accuracies than the SE training paradigm variants. Furthermore, overall the GFM-ALL approach obtains better results than the GFM-CAT approach. Most importantly, we see that the DeepPPMNet-DS-ALL variant outperforms state-of-the-art methods, such as ETS, BaggedETS, and ARIMA consistently across all performance metrics.

\begin{table}[!tb]
\caption{AO Dataset Results}
\centering
\resizebox{\columnwidth}{!}{%
\begin{tabular}{lcccc}\hline
	 		Method           				&Mean sMAPE	 		&Median sMAPE  		&Mean MASE 			&Median MASE\\ \hline
	 		DeepPPMNet-DS-ALL				&\textbf{0.1243}	&\textbf{0.1251}	&\textbf{0.9182}	&\textbf{0.8959}\\
	 		DeepPPMNet-DS					&0.1254				&0.1318		 		&0.9235				&0.9119\\
	 		Prophet							&0.1281				&0.1261				&0.9585  			&0.8964\\
	 		BaggedETS						&0.1286				&0.1299				&0.9472				&0.9466\\
	 		ETS								&0.1291				&0.1284				&0.9499				&0.9568\\
	 		ARIMA							&0.1332				&0.1351				&0.9691				&0.9762\\
	 		DeepPPMNet-SE-ALL				&0.1334				&0.1392		 		&0.9883				&0.9755\\
			DeepPPMNet-SE					&0.1417				&0.1435		 		&1.0340				&1.0289\\
			\hline

\end{tabular}
}
\label{tab:ao}
\end{table}

Table~\ref{tab:sa} summarises the results for the evaluations on the 88 LGAs of the SA category. It can be seen that the proposed DeepPPMNet-SE variant outperforms all other methods consistently across all error measures. Most importantly, we observe that contrary to our previous findings from the AO dataset, the SE variants achieve better results than the DS variants. Also, among the SE variants, we see that training a separate GFM gives better results than building one GFM across all the available time series. However, the observations among the DS based variants are reciprocal to those of SE.

\begin{table}[!tb]
\caption{SA Dataset Results}
\centering
\resizebox{\columnwidth}{!}{%
\begin{tabular}{lcccc}
\hline
	 		Method           				&Mean sMAPE	 		&Median sMAPE  	&Mean MASE 	&Median MASE\\ \hline
	 		DeepPPMNet-SE					&\bf 0.1128			&\bf 0.1126		&\bf 0.8817	&\bf 0.8088\\
	 		DeepPPMNet-SE-ALL				&0.1131				&0.1153			&0.8835		&0.8263\\
	 		ETS								&0.1146				&0.1201			&0.8893		&0.8212\\
	 		BaggedETS						&0.1148				&0.1193			&0.8996		&0.8299\\
	 		ARIMA							&0.1159				&0.1247			&0.8946		&0.8281\\
	 		Prophet							&0.1206				&0.1233	 		&1.0174  	&0.8988\\
	 		DeepPPMNet-DS-ALL				&0.1231				&0.1189		 	&0.9547		&0.8858\\
	 		DeepPPMNet-DS					&0.1266				&0.1210		 	&0.9652		&0.9164\\
			\hline

\end{tabular}
}
\label{tab:sa}
\end{table}

Table~\ref{tab:mo} shows the evaluation summary for the 87 LGAs of the MO category. We observe that the BaggesETS method achieves the best accuracy in terms of Mean sMAPE, and the DeepPPMNet-SE-ALL variant obtains the best Median sMAPE. Meanwhile, the DeepPPMNet-SE variant achieves the best Mean MASE, while the DeepPPMNet-SE-ALL variant and the Prophet method are the best-performing methods with respect to Median MASE.  Furthermore, similar to our previous findings from Table~\ref{tab:sa}, the DS variants show poor results compared to the SE variants, and overall the GFM-ALL approach obtains better results than its counterpart.

\begin{table}[!tb]
\caption{MO Dataset Results}
\centering
\resizebox{\columnwidth}{!}{%
\begin{tabular}{lcccc}
\hline

	 		Method           				&Mean sMAPE	 		&Median sMAPE  	&Mean MASE 	&Median MASE\\ \hline
	 		BaggedETS						&\bf 0.0917			&0.0859			&1.0751		&0.9946\\
	 		DeepPPMNet-SE-ALL				&0.0919				&\bf 0.0832		&1.0739		&\bf 0.9861\\
	 		DeepPPMNet-SE					&0.0927				&0.0919			&\bf 1.0703	&0.9946\\
	 		Prophet							&0.0929				&0.0951	 		&1.1264  	& \bf 0.9861\\
	 		ETS								&0.0930				&0.0912			&1.0861		&0.9946\\
	 		ARIMA							&0.0935				&0.0845			&1.0953		&0.9946\\
	 		DeepPPMNet-DS-ALL				&0.0973				&0.0983			&1.1195		&1.0703\\
	 		DeepPPMNet-DS					&0.1056				&0.1043			&1.1607		&1.1221\\
			\hline

\end{tabular}
}
\label{tab:mo}
\end{table}

Table~\ref{tab:ho} provides the results for the evaluations on the 80 LGAs of the HO category. Except for Median sMAPE, DeepPPMNet-SE-ALL outperforms all the benchmarks on each performance metric, whereas the proposed DeepPPMNet-DS achieves the best Median sMAPE. Similar to Table~\ref{tab:sa} and Table~\ref{tab:mo}, variants based on the SE training paradigm achieve better accuracies compared to those of DS. Overall, the GFM-ALL approach obtains better results compared to GFM-CAT approach.

\begin{table}[!tb]
\caption{HO Dataset Results}
\resizebox{\columnwidth}{!}{%
\begin{tabular}{lcccc}
\hline

	 		Method           				&Mean sMAPE	 		&Median sMAPE  	&Mean MASE 	&Median MASE\\ \hline
	 		DeepPPMNet-SE-ALL				&\bf 0.0780			&0.0634			&\bf 1.2118	&\bf 0.9685\\
	 		ETS								&0.0786				&0.0634			&1.2254		&0.9807\\
	 		BaggedETS						&0.0788				&0.0598			&1.2655		&0.9989\\
	 		DeepPPMNet-SE					&0.0804				&0.0634			&1.2137		&0.9925\\
	 		DeepPPMNet-DS-ALL				&0.0814				&0.0610			&1.2314		&1.0964\\
	 		ARIMA							&0.0809				&0.0654			&1.2811		&0.9989\\
	 		Prophet							&0.0813				&0.0647	 		&1.3476  	&1.0857\\
	 		DeepPPMNet-DS					&0.0825				&\bf 0.0597		&1.2433		&1.0964\\
			\hline
\end{tabular}
}
\label{tab:ho}
\end{table}

Table~\ref{tab:po} shows the results of the benchmarks for the 87 LGAs of the PO category. It can be seen that the proposed DeepPPMNet-SE variant achieves the best results on each performance metric. Once again, we observe that among the proposed training paradigms, the SE based variants perform better than the DS based variants. However, contrary to our previous findings, the DeepPPMNet variants that use the GFM-CAT approach, i.e., a separate DeepPPMNet is built only using the PO dataset, achieve better results compared to the GFM-ALL.  

\begin{table}[!tb]
\centering
\caption{PO Dataset Results}
\resizebox{\columnwidth}{!}{%
\begin{tabular}{lcccc}
\hline
	 		Method           				&Mean sMAPE	 		&Median sMAPE  	&Mean MASE 	&Median MASE\\ \hline
	 		DeepPPMNet-SE					&\bf 0.0620			&\bf 0.0528		&\bf 0.8606	&\bf 0.7759\\
	 		ETS								&0.0629				&0.0639			&0.8653		&0.8068\\
	 		BaggedETS						&0.0634				&0.0662			&0.8803		&0.8068\\
	 		DeepPPMNet-DS					&0.0669				&0.0661			&0.9366		&0.8539\\
	 		ARIMA							&0.0645				&0.0639			&0.8821		&0.8068\\
	 		DeepPPMNet-SE-ALL				&0.0664				&0.0664			&0.9071		&0.8068\\
	 		DeepPPMNet-DS-ALL				&0.0679				&0.0675			&0.9640		&0.8677\\
	 		Prophet							&0.0699				&0.0699	 		&1.0456  	&0.9297\\
			\hline

\end{tabular}
}
\label{tab:po}
\end{table}

In summary, we observe that the proposed DeepPPMNet variants achieve competitive results on the benchmark datasets, outperforming many state-of-the-art univariate forecasting techniques. This highlights the contribution made by employing a GFM strategy to exploit the similar EMS demand patterns. Moreover, we see that in the majority of cases, the GFM-ALL approach that uses all the available EMS category datasets performs better than the GFM-CAT approach that uses only the data specific to an EMS category. This can be mainly attributed to the low number of time series available in each EMS category, i.e., a maximum of 89 time series, for the AO category. On the other hand, grouping all the EMS categories together allows us to build a GFM across a total of 431 time series, and therewith allows our deep-learning based framework to learn better with more available data. It is also noteworthy to mention that the proposed DS training paradigm only outperforms the SE training paradigm in the AO dataset. This can be mainly attributed to the higher seasonal strength present in the AO dataset, compared to other EMS categories (see Fig.~\ref{lgademand2}).

\begin{figure*}[ht]
\centering
\subfloat[Type-1: 5\% ALI sensitivity]{
\includegraphics[width=0.50\textwidth,trim = 3cm 3cm 3cm 3cm]{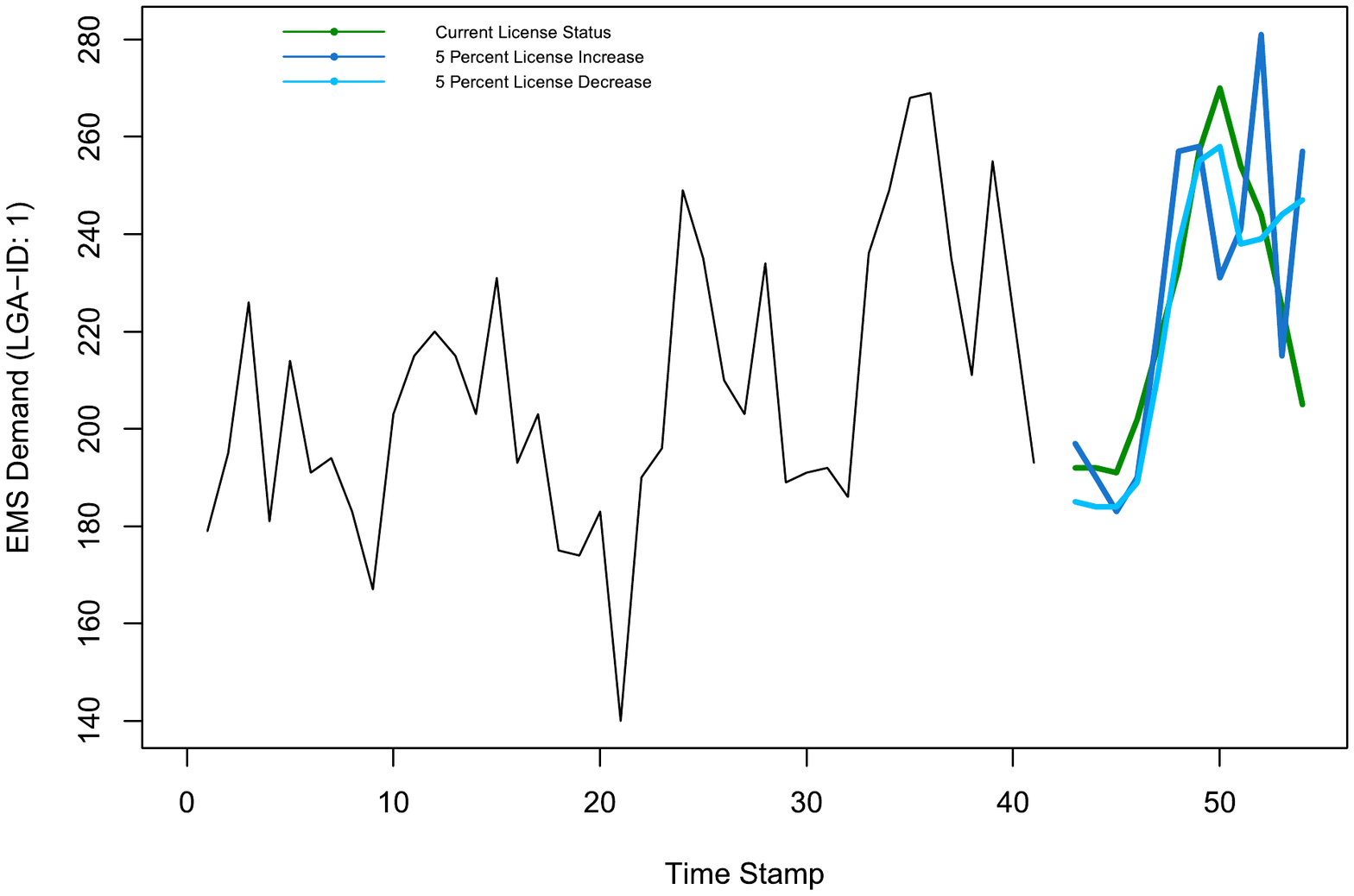}
\label{fig:subfig1}}
\subfloat[Type-2: 5\% ALI sensitivity]{
\includegraphics[width=0.50\textwidth,trim = 3cm 3cm 3cm 3cm]{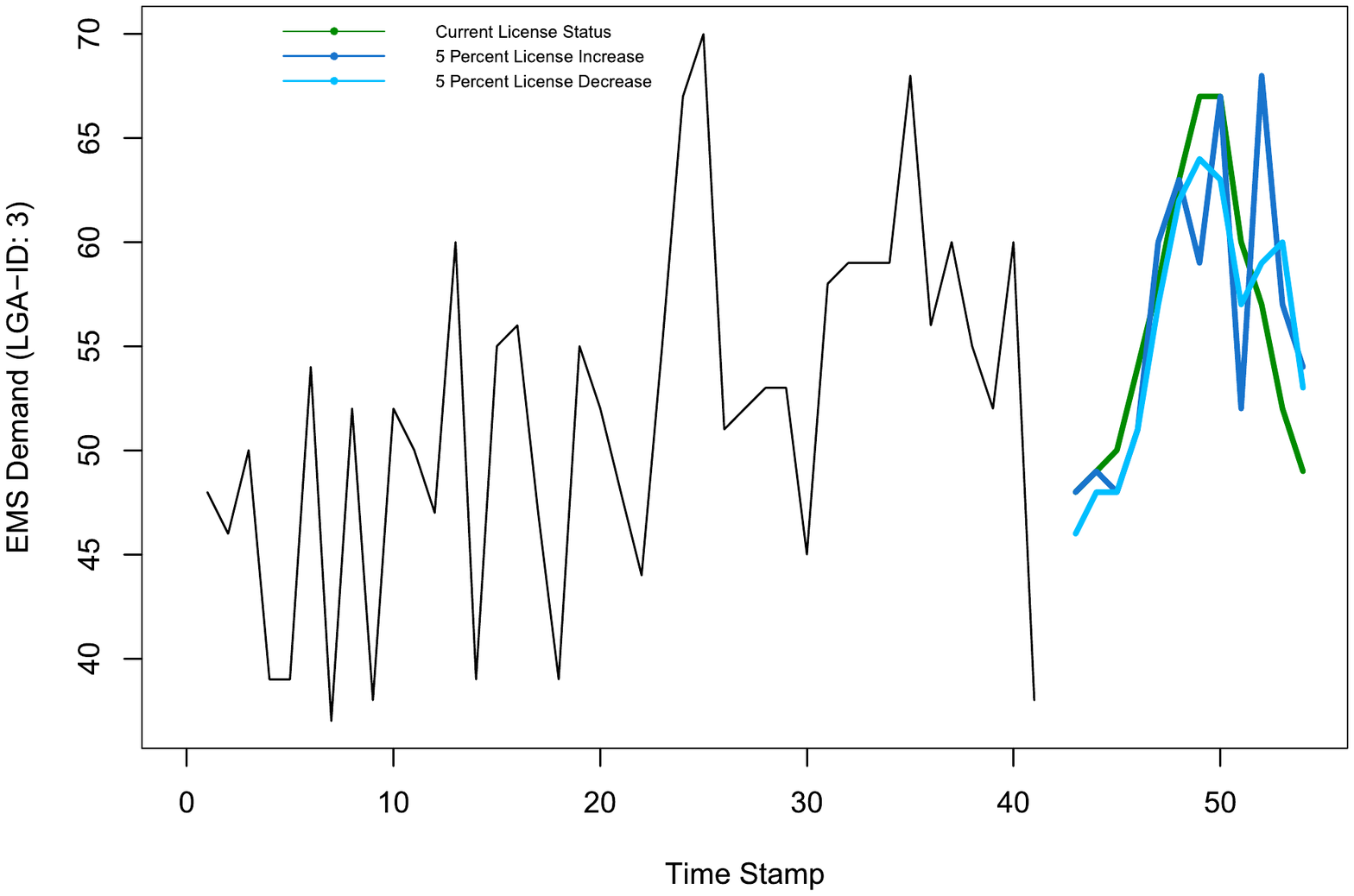}
\label{fig:subfig2}}
\qquad
\subfloat[Type-1: 10\% ALI sensitivity]{
\includegraphics[width=0.50\textwidth,trim = 3cm 3cm 3cm 3cm]{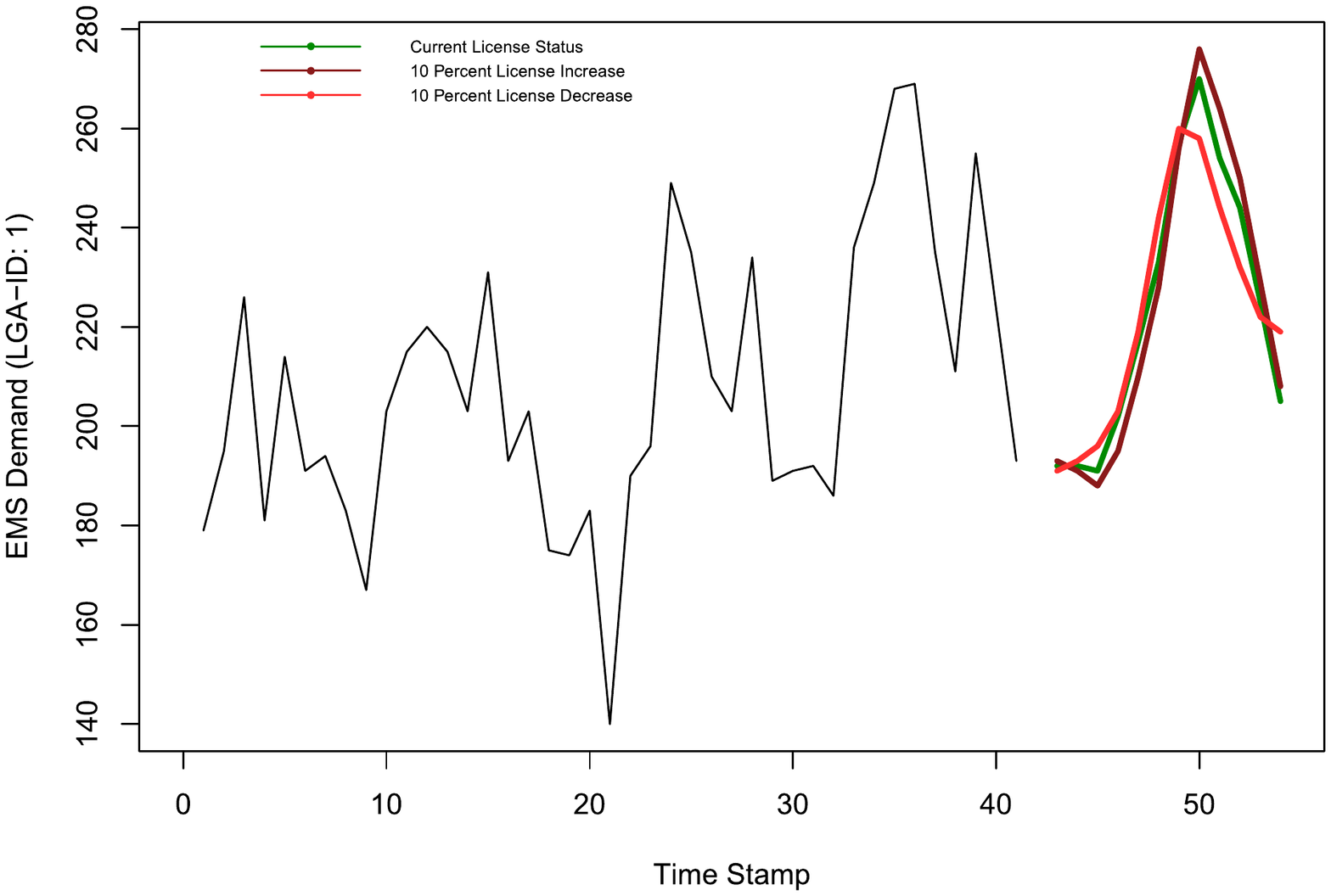}
\label{fig:subfig3}}
\subfloat[Type-2: 10\% ALI sensitivity]{
\includegraphics[width=0.50\textwidth,trim = 3cm 3cm 3cm 3cm]{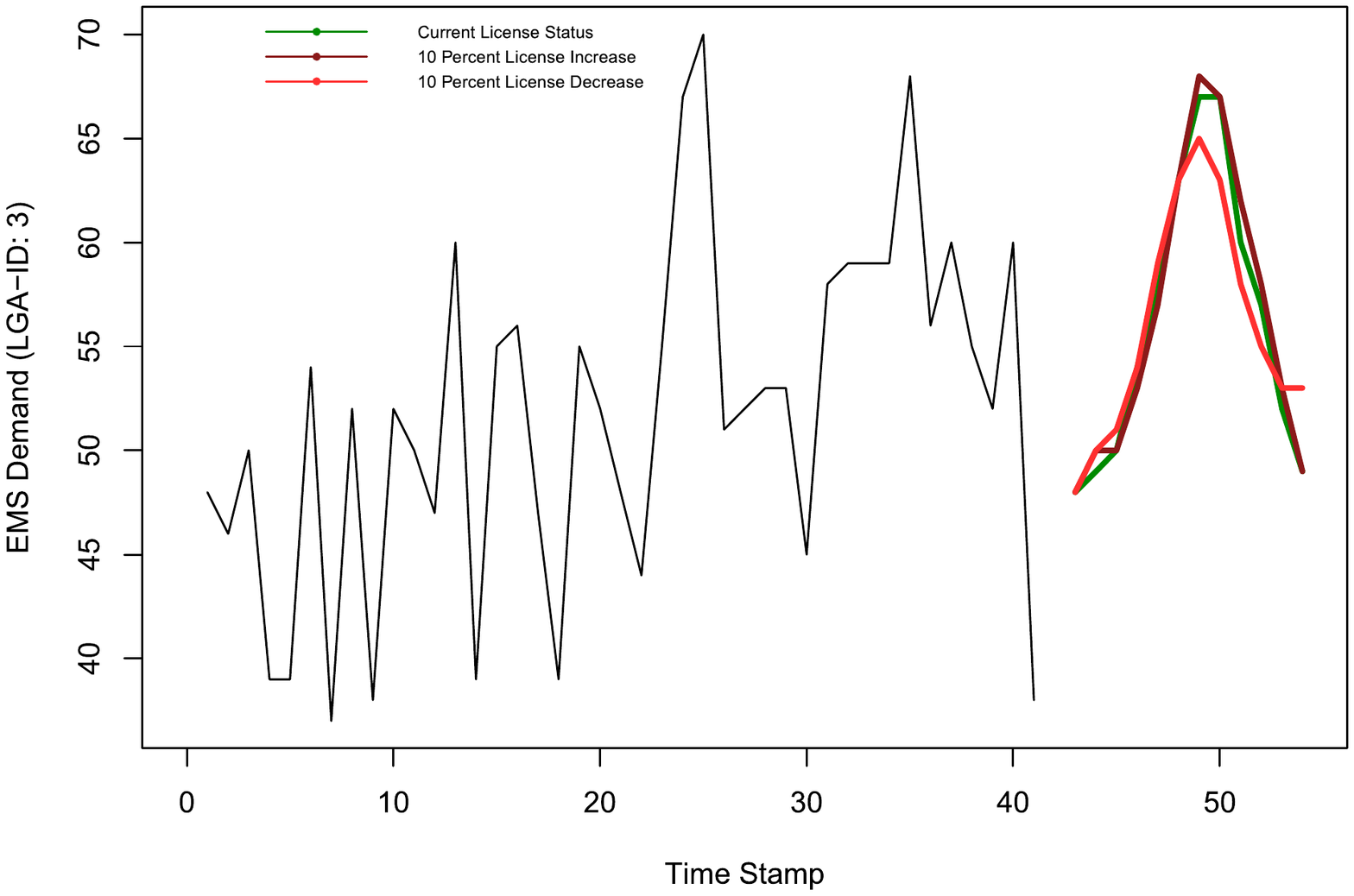}
\label{fig:subfig4}}
\caption{The application of `what-if' scenario analysis, using the number of  ALI (as a percentage of change) against the AO related EMS demand.}
\label{fig:sens}
\end{figure*}

\subsection{DeepPPMNet for Policy-making}

The competitiveness of the DeepPPMNet forecasting framework allows us to work towards assessing causing factors of EMS demand. In the following, we demonstrate the use of the DeepPPMNet framework as a platform to perform GC analyses using external time series. As a use case, we investigate how the number of alcohol licenses issued (ALI) for a certain period of time can affect the AO related EMS demand patterns. This can enable government policy makers to understand effects of different numbers of ALI by conducting `what-if' scenario analyses.

\subsubsection{Experimental Setup}

For this experiment, we use the monthly EMS data related to AO demand for 79 LGAs, and add the number of ALI as an exogenous variable to the current training input window. 
In particular, the DeepPPMNet uses a concatenation of the EMS demand and the ALI to learn the actual EMS demand of the output window.
We consider the five year period from January-2015 to May-2019, as for this period we have both the EMS data and the corresponding ALI data available. The observations of the last 12 months are used to test the DeepPPMNet, the rest of the data are used for training. We use the DeepPPMNet in the GFM-CAT mode, i.e., the GFM is built only using the alcohol related data, assuming other datasets are not available.  Also, in the decomposition layer, we replace the \textit{Prophet} method with \textit{STL}, as in this experiment the time series are shorter (42 monthly observiations), and \textit{STL} is more robust and better suited for shorter series. 
In this way, two additional DeepPPMNet variants are introduced, DeepPPMNet-SE-License and DeepPPMNet-DS-License, to represent the models that use the number of ALI as an exogenous variable.  
Furthermore, analogously we use the ARIMA benchmark in the exogenous-regressive mode, using the number of ALI as an additional regressor (ARIMAX-License).

\subsubsection{Accuracy Results}

Table~\ref{tab:aol} shows the results for the evaluations on the 79 LGAs of the AO category. 
Furthermore, we perform non-parametric Friedman rank-sum statistical significance testing to assess the statistical significance within the benchmarks, and the Hochberg’s post-hoc procedure is used to further examine these differences \cite{Garcia2010-jx}. The statistical testing is carried out on the sMAPE error measure, with a significance level of $\alpha$ = 0.05. 
Table~\ref{tab:alcoholstat} shows the corresponding results of the statistical testing evaluation. A horizontal line is used to separate the methods that perform significantly worse than the best performing method. The overall result of the Friedman rank sum test gives a $p$-value of $6.91 \times 10^{-11}$, which means the differences are highly significant. 

We see that DeepPPMNet-SE-License performs best, and achieves significantly better results than the DeepPPMNet-DS variants and the other benchmarks. However, we also observe that the DeepPPMNet-SE, BaggedETS, and ARIMAX-License variants do not perform significantly worse than the control method, DeepPPMNet-SE-License. 
The results confirm the earlier findings of the overall competitiveness of our methodology and the benefits of employing a GFM based forecasting model in this framework.

\subsubsection{Effect of Number of ALI}

Regarding the effect of the number of ALI, we see that after incorporating it as an exogenous variable (DeepPPMNet-SE-License), the accuracy of DeepPPMNet-SE has improved, outperforming the rest of the DeepPPMNet variants and the statistical benchmarks. We also observe that the ARIMAX-License variant has uplifted the results of the original ARIMA method, showing that inclusion of ALI as an exogenous variable is able to improve predictive accuracy over different methods.
Furthermore, we perform a non-parametric paired Wilcoxon signed rank test to explore the statistical significance of the differences. The test indicates a $p$-value of 0.687 for the DeepPPMNet-SE-License and DeepPPMNet-SE variants, and a $p$-value of 0.704 for the ARIMAX-License and ARIMA variants, which indicates that the differences are not statistically significant. 
%

As the results indicate that incorporating the number of ALI issued as an exogenous variable has improved the base accuracies of DeepPPMNet-SE and ARIMA, the number of ALI is a viable candidate for further exploration of a causal relationship. 
However, as the significance tests have not detected significant differences, further experiments with a larger dataset are needed in the future to clearly establish the relationship.

\subsubsection{Discussion of Causality of the Observed Effect}

As discussed in Section~\ref{sec:intro}, GC assesses an external regressor $Z_t$ with respect to other regressors $W_t$. We have established that the inclusion of ALI as $Z_t$ into our model is in fact able to improve forecasting accuracy across different methods. However, the strength of GC heavily depends on careful inclusion of additional factors $W_t$. In the current work, we suspect that ALI could work as a mere proxy to distinguish between rural and metropolitan areas, so, additional factors to be considered as $W_t$ could be, e.g., the amount of supermarkets in a given area, the amount of cafes or other non-alcohol-serving outlets, the amount of GPs, and others. The proposed DeepPPMNet framework is a platform where conveniently all these factors can be included in the model, to carefully assess the relationships. However, we leave such a careful analysis for future work, and illustrate next how to perform a series of `what-if' analyses on the AO related EMS demand once the GC analysis is done and the number of ALI is identified as a driving factor.

\begin{table}[!tb]
\centering
\caption{AO Results with Alcohol Licenses}
\resizebox{\columnwidth}{!}{%
\begin{tabular}{lcccc}
\hline

	 		Method           					&Mean sMAPE	 		&Median sMAPE  	&Mean MASE 	&Median MASE\\ \hline
	 		DeepPPMNet-SE-License				&\bf 0.1411			&\bf 0.1334		&\bf 0.9824	&\bf 0.9031\\
	 		DeepPPMNet-SE						&0.1424				&0.1414		 	&1.0009		&0.9964\\
	 		BaggedETS							&0.1476				&0.1403			&1.0501		&1.0000\\
	 		ARIMAX-License								&0.1494				&0.1485			&1.0472		&1.0116\\
	 		ARIMA								&0.1507				&0.1462			&1.0518		&0.9879\\
	 		ETS									&0.1505				&0.1447			&1.0636		&1.0121\\
	 		DeepPPMNet-DS						&0.1528				&0.1467		 	&1.0635		&0.9667\\
	 		DeepPPMNet-DS-License				&0.1539				&0.1478		 	&1.0703		&1.0175\\
	 		Prophet								&0.1870				&0.1862	 		&1.2522  	&1.2083\\
			\hline

\end{tabular}
}
\label{tab:aol}
\end{table}

\begin{table}[!tb]
\caption{Significance testing for AO Results}
\centering
\footnotesize
\begin{tabular}{ll}
	\toprule
	Method						&$p_{Hoch}$\\ \hline
	DeepPPMNet-SE-License 		&-\\
	DeepPPMNet-SE				&0.559 \\
	BaggedETS					&0.217 \\
	ARIMAX-License						&0.089\\
	\hline
	ETS							&0.038\\	
	ARIMA						&0.035\\	
	DeepPPMNet-DS				&0.006\\
	DeepPPMNet-DS-License		&1.729 $\times$ $10^{-4}$ \\
	Prophet						&2.254 $\times$ $10^{-15}$ \\
	\hline			
\end{tabular}
\label{tab:alcoholstat}
\end{table}

\subsubsection{`What-if' analysis}

Fig.~\ref{fig:sens} illustrates an example of performing a `what-if' scenario analysis to determine the sensitivity of ALI towards the AO specific EMS demand. Here, a 5\% increase/decrease (Fig.~\ref{fig:subfig1} and Fig.~\ref{fig:subfig2}) and a 10\% increase/decrease (Fig.~\ref{fig:subfig3} and Fig.~\ref{fig:subfig4}) is applied to the current number of ALI. We present two different types of demand variations present in two different LGAs. Fig.~\ref{fig:subfig1} represents the 5\% increase/decrease effect of the ALI on LGA-1, while Fig.~\ref{fig:subfig3} shows the 10\% increase/decrease effect of the ALI on LGA-1. We see that the increase of ALI has increased the predicted EMS demand, while a decrease in ALI has decreased EMS demand. Likewise, Fig.~\ref{fig:subfig2} and Fig.~\ref{fig:subfig4} show the 5\% increase/decrease and the 10\% increase/decrease effect of ALI on LGA-2. In this case, though decreasing the ALI has reduced the predicted EMS demand related to AO, the increase of ALI has not substantially affected the predicted EMS demand. Such insights may help government policy makers, when taking strategic decisions about the number of alcohol licenses to be issued for a specific LGA. Likewise, the DeepPPMNet can be used to conduct GC and `what-if' analyses on other EMS categories. For example, the number of pharmacy licenses issued could be explored as a causal factor that affects PO related EMS demand.

\section{Conclusion}
\label{sec:conclusion}

Overcoming the limitations of the current univariate state-of-the-art models, we have introduced DeepPPMNet, a three-layered, unified global forecasting framework that is capable of exploiting the non-trivial demand relationships present in EMS demand time series. 
We have evaluated DeepPPMNet using a real-world EMS dataset. The results have shown that the proposed DeepPPMNet framework has outperformed state-of-the-art univariate forecasting techniques, indicating the benefit of exploiting cross-series information available in the EMS demand time series. 

Furthermore, we have demonstrated how DeepPPMNet can be employed straightforwardly for causal analyses using the notion of Granger Causality, and how we subsequently can perform `what-if' scenario analyses to assist government authorities in the policy-making process.

As future work, we want to include further external factors in the analysis around alcohol outlet licenses, and we want to use DeepPPMNet on related cases, to demonstrate and make use of the full potential of our framework. From a methodologic point of view, we will enable the framework to perform probabilistic forecasting, for an even more informed policy making process.

\section{Acknowledgements}
\label{sec:ach}

This research was supported by the Australian Research Council under grant DE190100045, by a Facebook Statistics for Improving Insights and Decisions research award, and by Monash Institute of Medical Engineering seed funding. We acknowledge Sharon Matthews and the research assistants who code the NASS data and the ambulance services and paramedics who create and provide that data.

\bibliographystyle{IEEEtran}
\bibliography{reference}

\end{document}